\documentclass[sigconf]{acmart}

\AtBeginDocument{%
  \providecommand\BibTeX{{%
    \normalfont B\kern-0.5em{\scshape i\kern-0.25em b}\kern-0.8em\TeX}}}

\setcopyright{acmcopyright}
\copyrightyear{2018}
\acmYear{2020}
\acmDOI{10.1145/1122445.1122456}

\acmConference[KDD '20]{KDD '20}{Aug 22--27, 2020}{San Diego, CA}
\acmPrice{15.00}
\acmISBN{978-1-4503-XXXX-X/18/06}

\acmSubmissionID{2042}
\usepackage[]{algorithm2e}
\usepackage{multirow}

\begin{document}

\title{Variable Name Recovery in Decompiled Binary Code \\ using Constrained Masked Language Modeling}

\author{Pratyay Banerjee}
\email{pbanerj6@asu.edu}
\author{Kuntal Kumar Pal}
\email{kkpal@asu.edu}
\affiliation{%
  \institution{Arizona State University}
  \city{Tempe}
  \state{Arizona}
  \postcode{85281}
  \country{USA}
}

\author{Fish Wang}
\email{fishw@asu.edu}
\author{Chitta Baral}
\email{chitta@asu.edu}
\affiliation{%
  \institution{Arizona State University}
  \city{Tempe}
  \state{Arizona}
  \postcode{85281}
  \country{USA}
}


\begin{abstract}
Decompilation is the procedure of transforming binary programs into a high-level representation, such as source code, for human analysts to examine.
While modern decompilers can reconstruct and recover much information that is discarded during compilation, inferring variable names is still extremely difficult.
Inspired by recent advances in natural language processing, we propose a novel solution to infer variable names in decompiled code based on Masked Language Modeling, Byte-Pair Encoding, and neural architectures such as Transformers and BERT.
Our solution takes \textit{raw} decompiler output, the less semantically meaningful code, as input, and enriches it using our proposed \textit{finetuning} technique, Constrained Masked Language Modeling. Using Constrained Masked Language Modeling introduces the challenge of predicting the number of masked tokens for the original variable name.
We address this \textit{count of token prediction} challenge with our post-processing algorithm.
Compared to the state-of-the-art approaches, our trained VarBERT model is simpler and of much better performance.
We evaluated our model on an existing large-scale data set with 164,632 binaries and showed that it can predict variable names identical to the ones present in the original source code up to 84.15\% of the time.
\end{abstract}

\ccsdesc[500]{Security and privacy~Software reverse engineering}
\ccsdesc[500]{Computing methodologies~Natural language processing}
\ccsdesc[500]{Computing methodologies~Machine learning}
\ccsdesc[500]{Computing methodologies~Model development and analysis}

\keywords{reverse engineering, decompilation, variable name recovery, masked language modeling, bert}



\maketitle

\section{Introduction}

Despite the recent progress in the software open source movement, source code is oftentimes unavailable to security researchers, analysts, and even software vendors.
This makes \emph{software reverse engineering}, a family of techniques that aim at understanding the behavior of software without accessing its original source code, irreplaceable~\cite{david2019neural}.
Reverse engineering techniques are essential to accomplishing a set of security-critical tasks, including malware analysis, vulnerability discovery, software defect diagnosis, and software repairing~\cite{allamanis2018survey}.

With the rapid development of modern decompilers (e.g., Hex-Rays Decompiler~\cite{chagnonida, hex2013hex} and Ghidra~\cite{ghidra}), decompilation, or reverse compilation, is gradually becoming an essential technique for software reverse engineering tasks.
Built atop advanced program analysis techniques and sophisticated heuristics, these decompilers can identify much information that is lost during compilation and binary stripping, such as function boundaries, function prototypes, variable locations, variable types, etc.
Names of variables in a program usually embed much semantic information that is crucial for understanding the behaviors and logic of the program.
Nevertheless, the state-of-the-art decompilers are all limited to recovering structural information and cannot recover meaningful variable names.
Therefore, the usual first step taken by a decompiler user is painfully interpreting the decompiled code and renaming the unnamed variables one by one.
This is, unfortunately, an insurmountable obstacle for reverse engineering stripped binary programs.


While researchers have taken steps in predicting variable names in high-level programming languages~\cite{jaffe2018meaningful, raychev2015predicting, vasilescu2017recovering, allamanis2014learning, bavishi2018context2name, artuso2019function}, it is worth noting that inferring variable names in decompiled binary code poses a unique set of challenges.
High-level programming languages like Java, Python, and JavaScript are syntactically rich:
Variable types are preserved in these languages, while they are usually eliminated in binaries.
In fact, type inferencing on binary programs still remains an open problem~\cite{Caballero2016}.
Moreover, modern compilers that generate machine code produce all sorts of irreversible changes~\cite{Yakdan2015}, which diminish the power of variable name prediction techniques that are based on local data dependencies.
DEBIN~\cite{he2018debin} and DIRE~\cite{lacomis2019dire} are the pioneers in predicting variable names in stripped binary programs.
DEBIN proposes a statistical model that yields poor performance and generalizability.
DIRE produces much better results by employing an LSTM\cite{hochreiter1997long} and Gated Graph Neural Network-based model~\cite{li2015gated} that is trained on a huge corpus of source code with sophisticated structural features extracted from Abstract Syntax Trees.
Their approach is complex and error-prone.


Recently, there has been a substantial improvement in the field of natural language processing with the advent of techniques such as Transformers~\cite{vaswani2017attention}, Masked Language Modeling~\cite{devlin2018bert}, and BERT~\cite{devlin2018bert}.
With the help of these techniques, we propose a novel solution to predicting variable names in stripped binary programs using only the \textit{raw} decompiled code as input.
We propose \textit{Constrained Masked Language Modeling}, a modification of the Masked Language Modeling task to train a BERT model to recover variable names.
We learn a vocabulary of code-tokens using Byte-Pair Encoding and learn rich contextual representation using Masked Language Modeling.


Using Masked Language Modeling to predict variable names introduces a new challenge of predicting the \textit{Count of Variable Name Tokens}, i.e., determining how many mask tokens are required to predict a variable name.
We address this challenge by using a heuristics-based algorithm.

Both DIRE and our aforementioned solutions are computationally expensive.
After careful analysis of the nature of the task, we argue that predicting variable names in the decompiled code does not require such a deep and computationally intensive neural architecture.
To support our argument, we train a smaller BERT model and achieve a similar performance;
Additionally, it is capable of taking longer input sequences of code.

We train and evaluate our models using the DIRE data set, a collection of 164,632 x86-64 binary programs generated from C projects crawled from GitHub~\cite{lacomis2019dire}.
Our evaluation results show that our models can predict variable names that are identical to the ones used in original source code up to 84.15\% of the time, which surpasses our baselines, DEBIN and DIRE.

\smallskip

\noindent
Our contributions are summarized below:
\begin{itemize}
    \item We adapt Masked Language Modeling for the task of variable name recovery in decompiled code and propose Constrained Masked Language Modeling.
    \item We address the challenge of \emph{Count of Variable Name Tokens} using a heuristic-based algorithm and evaluate its performance.
    \item We train two BERT models, VarBERT-base and VarBERT-small, that only use \emph{raw} decompiled code as input.
    Models and code will be open-sourced.
    \item In the task of variable name recovery, we surpass DIRE, the state-of-the-art solution, on a large data set of 164,632 binary programs.
    We improve by \textbf{12.36}\% in overall accuracy and \textbf{49}\% in generalizability, that is, our model can successfully predict variables names that are not present in the training set.
\end{itemize}

In the spirit of open science, We will open source our source code and trained BERT models upon the publication of this paper.

\section{Background and Related Work}

Our solution leverages techniques in machine learning and natural language processing to refine the results of binary code decompilation.
Before diving into the details of our solution, in this section, we will first present the necessary background as well as some work that is closely related to our proposed solution.

\subsection{Binary Code Decompilation}

Binary code decompilation is the process of converting compiled binary code to a higher-level representation, oftentimes C source code or C-like pseudocode.
Compilation and binary stripping are lossy procedures where much semantic information is discarded since it is useless to CPUs.
Such information, such as control flow structures, function names, function prototypes, variable types, and variable names, is very important to software reverse engineering tasks.
While modern decompilers have made significant progress in recovering and inferring various types of lost information, recovering variable types in decompiled code remains challenging.

\subsection{NLP and Program Analysis}

Structured programs and natural languages have similar statistical properties~\cite{allamanis2018survey, hindle2012naturalness, devanbu2015new}.
This has led to the application of statistical models in program analysis and software engineering.
Some of the notable applications are code completion~\cite{bruch2009learning, proksch2015intelligent, omar2013structured}, code synthesis~\cite{rabinovich2017abstract, beltramelli2018pix2code, gvero2015synthesizing, ling2016latent, kushman2013using, yin2017syntactic}, obfuscation~\cite{liu2016towards, pu2016sk_p}, code fixing~\cite{gupta2018deep, gupta2017deepfix}, bug detection~\cite{ray2016naturalness, wang2016bugram}, information extraction~\cite{cerulo2013hidden, sharma2015nirmal, yadid2016extracting}, syntax error detection \cite{campbell2014syntax} and correction \cite{bhatia2016automated}, summarization~\cite{hu2017codesum,iyer2016summarizing} and clone detection~\cite{white2016deep}.


\subsection{Statistically Predicting Variable Names}

The probabilistic graphical model, Conditional Random Fields (CRF) has been useful in the prediction of syntactic names of identifiers and their semantic type information of JavaScript programs. Their system JSNICE \cite{raychev2015predicting} were able to correctly predict 63\% of the names and 81\% of the type information. Though our goal of prediction of variable names is same our approach and program domain differ considerably from theirs.

Another system A\textsc{UTONYM} \cite{vasilescu2017recovering} surpassed JSNICE in the prediction of variable names for Javascript codes using statistical machine translation. Jaffe et al. \cite{jaffe2018meaningful} also used statistical machine translation to generate meaningful names to the variables for the decompiled source codes written in C. They could achieve 16.2\% accuracy in predicting meaningful names in a dataset of size 1.2TB of decompiled C code. They predicted variable names that are similar and convey the same meaning as the original names, whereas, we try to predict the original variable names from the decompiled code. In that respect, we keep a much stricter evaluation metric.

The NATURALIZE~\cite{allamanis2014learning} framework has been quite successful (94\% accuracy) in suggesting meaningful identifier names and format styles using n-gram probabilistic models.
However, we use different models and training methods for the task.


With the help of machine learning techniques, researchers have made much progress in recovering variable names on decompiled binary code.
However, existing techniques suffer from some major drawbacks.
DEBIN uses probabilistic models like CRF~\cite{he2018debin};
Unfortunately, as shown in another paper, its accuracy and generalizability are poor, which renders DEBIN nearly useless in real-world settings~\cite{lacomis2019dire}.
DIRE leverages neural network on Abstract Syntax Trees (ASTs) extracted from collected source code~\cite{lacomis2019dire};
While DIRE yields a good prediction performance, the AST extraction process can be tedious and error-prone.

In this paper, we show that with advanced NLP techniques, it is possible to achieve a significantly better performance in accuracy and generalizability of variable name prediction in decompiled binary code.
Moreover, our proposed solution does not require extracting ASTs from decompiled code.
Instead, it directly uses \emph{raw} decompiled code as input, which is simpler and more robust than DIRE.

\subsection{Neural Models}

The success of statistical models and progress in the development of stronger neural models led to the application of neural models for identifier name recovery. Our work is related to the following recent works in this area. 

In a recent work, \textsc{CONTEXT}2N\textsc{AME} \cite{bavishi2018context2name} attempted to assign meaningful names to the identifiers based on the context of minified JavaScript codes. They were able to successfully predict 47.5\% of meaningful identifiers on 15,000 minified codes using recurrent neural networks. Our work differs from theirs in the sense that we use much advanced deep learning models, we predict the actual original names instead of assigning similar meaningful names and we predict variables for decompiled C code instead of minified JavaScript codes.

Few of the works attempted to predict function names from stripped binaries. Artuso et al. \cite{artuso2019function} used sequence to sequence networks in two settings (with or without pre-trained embeddings) to predict function names in a dataset created using stripped binaries compiled from 22,040 packages of Ubuntu apt repository with the precision and recall of 0.23 and 0.25 respectively. In another work, the NERO model by David et al. \cite{david2019neural} predicted the procedure names in a dataset created from Intel 64-bit executables running on Linux, using LSTM  with a precision and recall of 45.82 and 36.40 respectively. Our approach differs from both of the works in the sense that we focus on predicting the variable names instead of functions  and our approach to the task.

DIRE \cite{lacomis2019dire} is the most recent work related to us. They were able to successfully predict the original variable names 74.3\% of the time in the decompiled source code of 164,632 unique x86-64 C binaries mined from Github. They used gated graph neural networks(GGNN) \cite{li2015gated} and bidirectional LSTM to encode the Abstract Syntax Tree (structural information) and the decompiled code(lexical information) followed by a decoder network with attention. For our work we use the same dataset published by them but our training method and inputs differ considerably.

\subsection{BERT}
Bidirectional Encoder Representations from Transformers is \textit{designed to pre-train deep bidirectional representations from the unlabeled text by jointly conditioning on both
left and right context in all layers} \cite{devlin2018bert}. BERT is pre-trained in an unsupervised way on a huge collection of natural text for two tasks, first, the masked language modeling (MLM) and second, the next sentence prediction (NSP) to make the model understand the relation between tokens and sentences respectively. Since it's release it has become ubiquitous in almost all tasks in various domains.

\begin{figure}
  \includegraphics[width=8.5 cm]{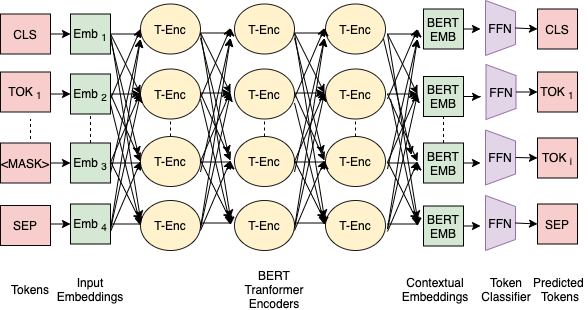}
  \caption{BERT Mask Language Modeling}
  \Description{BERT Masked Language Modeling}
  \label{fig:bert}
\end{figure}


\subsection{Vocabulary for Neural Models}
All the neural models make use of a specific vocabulary set created from the training data to translate words or tokens to numeric encodings. It is  created using a tokenizer on the natural language texts. The size of the vocabulary should not be too low to cover most of the words in training, validation and test data. It should not be too high either to create a massive vocabulary, which may lead to considerably large learned vector embeddings.

\subsubsection{WordPiece and SentencePiece Embedding} To keep an optimal size of the vocabulary the WordPiece embedding \cite{wu2016google} was introduced. In this embedding, each word is divided into a limited set of common sub-units(sub-words). The word-piece embedding is helpful in handling the rare words in the test samples. BERT uses WordPiece embeddings with a vocabulary of size 30,000. Unknown words are split into smaller units which are present in the vocabulary. Another type of embedding \cite{kudo2018sentencepiece} can be directly trained from raw sentences without the need for pre-segmentation of text as compared to earlier approaches. It helps user to create a purely end-to-end and language-independent system. 

\subsubsection{Byte-Pair Encoding}
Byte-Pair Encoding(BPE) \cite{sennrich2015neural} is a hybrid between the character and word level text representation. It can handle large vocabulary. The full word is broken down into smaller sub-unit words after performing statistical analysis on the training data. Another implementation of BPE \cite{radford2019language} uses bytes instead of unicode characters as base sub-unit words.

\section{Approach Overview}
\begin{figure}
  \includegraphics[width=8.5 cm]{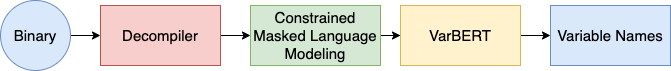}
  \caption{Overall End-to-end Approach}
  \Description{Overall End-to-end Approach}
  \label{fig:e2e}
\end{figure}

Our approach involves the following. We extract the decompiled \textit{raw} code from the binaries. Each function is taken as an independent input instance. We create a corpus by using all such \textit{raw} code to learn a vocabulary of most frequent code-tokens using the technique of Byte-Pair Encoding. We then proceed to learn representations of the code-tokens and a BERT model using the pre-training method of Masked Language Modeling. Finally, we fine-tune the BERT model using our Constrained Masked Language Modeling technique, to predict only the variable names.

\section{Dataset Description}

In this work, we use the DIRE\cite{lacomis2019dire} dataset. The dataset consists of 3,195,962 decompiled x86-64 functions and
their corresponding abstract syntax trees with proper annotations of the decompiled variable name and their corresponding original gold-standard names. The dataset was created by scraping open-source C codes from Github. The codes have been compiled keeping the debug information and then decompiled using Hex-Rays \cite{hex2013hex}, a state-of-the-art industry decompiler. The decompiled variable names are the names assigned to the original variable names by the Hex-Rays. 
We worked on the reduced preprocessed dataset released by the authors with the same train-dev-test(80:10:10) splits. The number of train, validation and test files is 1,00,632, 12,579 and 12,579 respectively.
In the dataset, there are 1,24,702, 1,24,179, and 10,11,054 functions which involve 6,74,854, 6,80,493, and 55,23,045 variables in validation, test and train files respectively.  

\subsection{Vocabulary}
We use Byte-Pair Encoding to learn the vocabulary. We generate a corpus by first replacing all decompiler generated variables in the \textit{raw} code with original variable names and then concatenating all decompiled functions. We use the HuggingFace Tokenizers tool to learn the Byte-Pair Encoding vocabulary. We learn two sets of vocabulary, one with 20,000 tokens, and another with 50,000 tokens to compare the results. Both are generated with allowed merges of 50,000. 

\section{Model Description}
\subsection{Masked Language Modeling}

\begin{figure*}
  \includegraphics[width=17 cm]{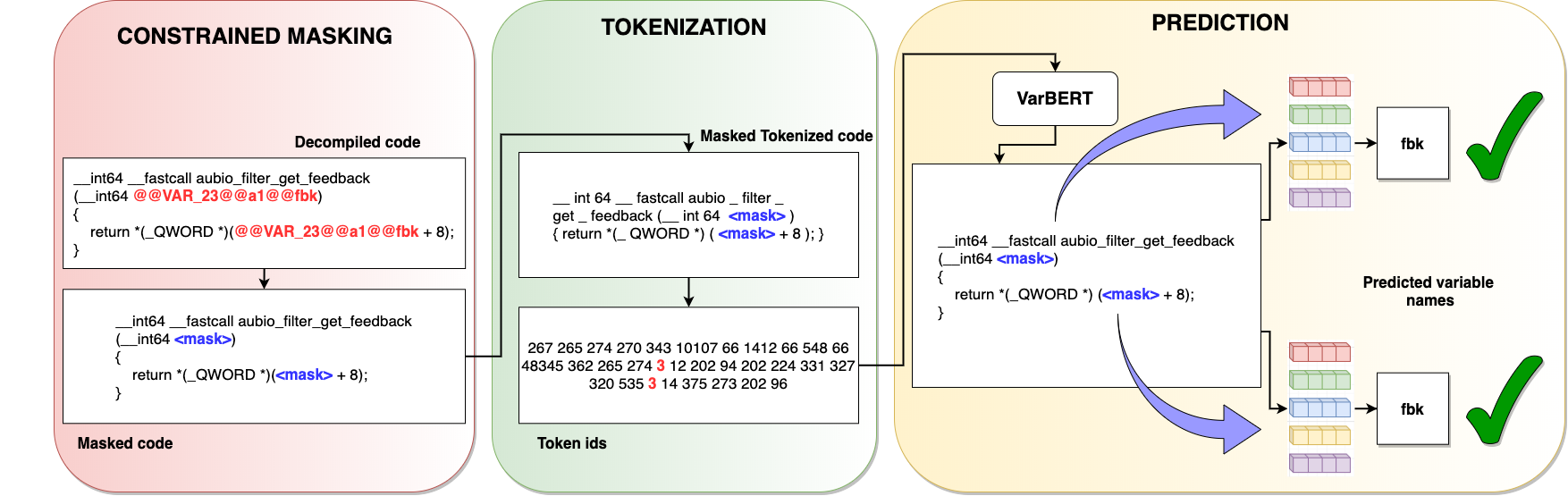}
  \caption{End-to-end Constrained Masked Language Modeling}
  \Description{End-to-end Constrained Masked Language Modeling}
  \label{fig:bertmlm}
\end{figure*}

Traditional Language Modeling is the task of predicting the next word or token given the previous set of words or tokens. Formally, the task of Language modeling can be represented as learning the following probability for all words in a given vocabulary:
\begin{equation}
 P(W_i=V_j | {W_{i-1},..,W_0})
\end{equation}
, where $W_i$ are the word or tokens, ${W_{i-1},..,W_0}$ is the prior sequence, and $V_j$ are the tokens present in the vocabulary.

In \cite{devlin2018bert}, they proposed a different language modeling task  to train Transformer Encoders to learn rich representations for tokens. This task, Masked Language Modeling, is defined as follows. Let $W_0,....W_N$ be a sequence of tokens. A random set of $m$ tokens from this sequence is chosen and replaced with a special mask token, $[mask]$. Then we learn the following probability for each of the masked tokens:
\begin{equation}
 P([mask]_i=V_j| {W_0,..,[mask]_k..,W_N})
\end{equation}
, where $[mask]_i$ represents the $i^{th}$ masked token, $V_j$ represents the tokens in the vocabulary set $V$, $W_0,..,[mask]_k..,W_N$ is the sequence where $m$ random tokens are masked and $[mask]_k$ is different masked token. 

In Masked Language Modeling, not all tokens which are chosen to be replaced with $[mask]$ are  replaced in the input. We choose 15\% of the token positions at random for
prediction. If a token is selected, we replace 80\% of selected tokens with $[mask]$, 10\% of the tokens with another random token, and the rest 10\% are kept unchanged. The set of masked tokens can change for a particular instance of data across multiple epochs. This is called \textit{dynamic masking}.
The model is trained to predict the original token with a cross-entropy loss.

It has been demonstrated in several applications the impact of such a modeling task in learning-rich token vector representations using unsupervised or self-supervised learning methods \cite{devlin2018bert,liu2019roberta}. 

\subsection{Whole-Word Masking}
In Masked Language Modeling, since we use a random sampling of tokens to mask, we may mask sub-parts of a word. As seen in the example, a big word that is not present in the vocabulary can be split into multiple tokens. A subset of these tokens may be chosen to be masked. To learn better representations, and ensure the model captures the entire word, we perform Whole-Word Masking, in which if a word is chosen to be masked, all the tokens of the word are masked and the model is trained to predict all the original token.

\subsection{Constrained Masked Language Modeling}

Constrained Masked Language Modeling is a variation of Masked Language Modeling where the tokens are not masked at random. We define Constrained Masked Language Modeling as follows:

Let $W_0,....W_N$ be a sequence of tokens. Let $C=\{A_0,..A_c\}$ be a set of tokens which we define as the constrained set of tokens. Then in constrained masked language modeling, all tokens in $W_0,....W_N$ which belong to $C$ are masked. $C$ is a subset of $V$. Here all the tokens are replaced with a masked token, unlike Masked Language Modeling. Similar to Masked Language Modeling, we train the model to predict the masked token with a cross-entropy loss:

\begin{align}
    -\sum_{i=0}^N\sum_{c=1}^My_{w_i,v_c}\log(p_{w_i,v_c})
\end{align}
, where $M$ is the size of the vocabulary, $w_i$ is the current token, $v_c$ is the target token, $y$ is an indicator variable which is 1 if the target is $v_c$ and 0 otherwise and $p$ is the probability $w_i$ is same as $v_c$.

\subsection{Count of Token Prediction}
In the task of Variable Name Recovery, during test time, we do not know the number of tokens the original variable should have.
We define the Count of Token Prediction as the task to determine the count of mask tokens during test time.
We solve this with the following heuristic defined in Algorithm \ref{algo1}. We evaluate our BERT models with both an Oracle model which gives us the number of tokens and our heuristics. The max allowed number of tokens is derived from a statistical analysis of the different variable names present in the Train dataset.

\begin{algorithm}[]
\SetAlgoLined
\KwResult{BestPredictedVariable}
 Count=1\;
 MaxProbability=0\;
 BestPredictedVariable=""\;
 MaxAllowedToken\;
 Model\;
 \While{Count<MaxAllowedToken}{
  Place count number of mask tokens\;
  CurrVariable=Model.Predict\;
  P=Mean of the probabilities for each token\;
  \If{P>MaxProbability}{
    BestVariable=CurrVariable\;
    MaxProbability=P\;
    }
 }
 \caption{Best Variable Selection Heuristics}
\label{algo1}
\end{algorithm}

\subsection{Technical Details}
A Transformer architecture \cite{vaswani2017attention} contains multiple layers, with each block using multiple self-attention heads. We train two different versions of VarBERT each differing in the number of layers $L$, the number of self-attention heads $A$ and the hidden dimension $H$. The vocabulary for both versions is the same. Training such a huge architecture requires a lot of computing and training time. We define a fixed hyper-parameter budget of two training runs over the entire dataset for each of the versions to limit our training time and compute cost. We take initial hyper-parameters from RoBERTa \cite{liu2019roberta}.

\subsubsection{VarBERT-Base}
This version has $L=12$, $A=12$ and $H=768$. This leads to total parameters to be trained to be nearly 125 million. The maximum number of input tokens this version can take is 512.

\subsubsection{VarBERT-Small}
We hypothesize in the task of Variable Name Recovery, a smaller model in terms of depth and parameters might perform reasonably well, and hence create this version which has $L=6$, $A=8$ and $H=512$. We though increase the number of input tokens the model can take to 1024 to be able to train the model with longer code sequences and be more useful for the community. The number of trainable parameters for this model is 45 million, which is 2.5 times smaller than VarBERT-Base.

\subsection{Training Methodology}
For the task of Variable Name Recovery, we train our models with the following methodologies. 
\subsubsection{Pre-Training}
The need and impact of Pre-Training on auxiliary tasks before the final intended task has been demonstrated in multiple previous natural language processing and computer vision work such as in \cite{devlin2018bert,liu2019roberta,yang2019xlnet,tan2019lxmert}. For the task of Variable Name Recovery, we define the Masked Language Modeling task as the pre-training task. We follow the same steps as done in RoBERTa \cite{liu2019roberta} to train our model and learn rich representations for our code-tokens. We train, both with and without Whole-word Masking. 

\subsubsection{Finetuning}
We finetune our models, for the Variable Name Recovery task using the Constrained Language Modeling task. 

\subsubsection{Optimization}
We optimize our models using BERTAdam \cite{devlin2018bert,kingma2014adam} with following parameters: $\beta_1=0.9$ and $\beta_2=0.999$, $\epsilon=1e-6$ and $L_2$ weight decay of 0.01. We warmup over first 10,000 steps to a peak value of $1e-4$ and then linearly decay. We set our dropout to 0.1 on all layers and attention weights. Our activation function is GELU \cite{hendrycks2016gaussian}.

We use HuggingFace Transformers and Facebook FairSeq tools to train our models.

\section{Experiments}
\subsection{Experimental Setup}
Our models are evaluated on the DIRE dataset. Split of the dataset given in DIRE \cite{lacomis2019dire} is 80:10:10. We use only the \textit{raw} code generated from the decompilers and replace the decompiler generated variable names with the original variable name given by the developer. We evaluate the impact of pre-training and whole-word masking. We also evaluate the impact of constrained masked language modeling using the pre-trained only model as a baseline. We evaluate the impact of the size of the vocabulary and the size of the models.  

As defined above, we train the models with a hyper-parameter budget of two, so a bigger budget might result in even better performance. The hyperparameters for pre-training and finetuning are a batch size of 1024 and the number of epochs is restricted to 40.
We use 4 Nvidia Volta V-100 16GB graphics cards to train our models.
Approximately, the VarBERT-Base has a training time of 72 hours and VarBERT-Small has a training time of 38 hours. 

\subsection{Metrics}

We evaluate our models using the following metrics. \textit{Exact Match accuracy} of predicting the correct variable name. The final goal of such a model is to provide suggestions to system reverse engineers, and our models also provide a ranking of tokens, we measure the accuracy of the correct variable at different ranks, which are 1,3,5 and 10. In DIRE \cite{lacomis2019dire} they also measure the character error rate (CER) metric, which calculates the edit distance between the original and predicted names, then normalizes the length of the original name, as defined in CER \cite{wang2016character}. We measure this for our heuristic-driven Count of Token Prediction algorithm. We also measure the \textbf{Perplexity} of both language modeling tasks. Perplexity measures how well a probabilistic language model predicts a target token.


The binaries in the dataset use C libraries. The different splits Train, Validation and Test contain binaries that share these libraries. The functions in these libraries have the same variable names and bodies. To better understand the generalizability of our proposed models and techniques, we also report the metrics on two sets, \textit{Body not in the train} and \textit{Overall}. These sets are defined as meta-tags in the DIRE dataset.

\subsection{Baselines}
Our baseline models for the Variable Name Recovery tasks are the following.

\subsubsection{DEBIN}
It uses statistical models such as CRFs and Extremely Randomized Trees with several handcrafted features to recover variable names, along with other debugging information from stripped binaries. Few of the handcrafted features are functions used, registers used, types, flags, instructions and relationships between functions, variables, and types. This acts as a weak baseline for our proposed models.

\subsubsection{DIRE}
It uses LSTMs, Gated Graph Neural Networks and Attention Mechanism with an Encoder-Decoder architecture to generate variable names. It also uses the \textit{raw} code as one of the inputs. In addition to the decompiled code, it uses a GNN based structural encoder to encode Abstract Syntax Trees. This acts as a strong neural baseline for our proposed models. 

Both the above models, use the \textit{raw} code to recover variable names but also use additional features. In our proposed models, we only use the \textit{raw} code.

\section{Discussion and Analysis}
\subsection{Dataset Analysis}
\begin{figure}
\small
  \includegraphics[width=7 cm]{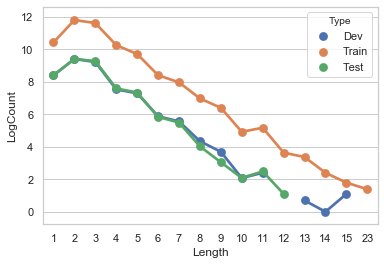}
  \caption{Distribution of length of gold tokenized variable names in Train, Validation and Test data. The counts are in Natural Log Scale.}
  \Description{Distribution of length of gold tokenized variable names in Train, Validation and Test data}
  \label{fig:goldvar}
\end{figure}

\begin{figure}
\small
  \includegraphics[width=7 cm]{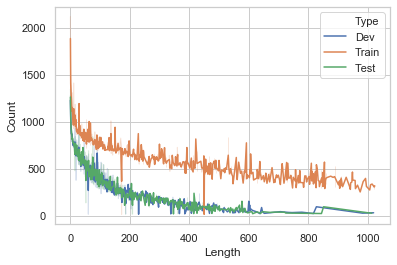}
  \caption{Distribution of length of function in Train, Validation and Test data}
  \Description{Distribution of length of function names in Train, Validation and Test data}
  \label{fig:funclen}
\end{figure}

We start with the initial analysis of the dataset. We measure the overlap between the variables present in the Train, Validation and Test set splits with our learned vocabulary. We also measure the length of the functions. Figure \ref{fig:goldvar} and Figure \ref{fig:funclen} show the respective distributions. 216 and 167 functions from the Validation and Test set were truncated due to the limitation of maximum allowed sequence length of 1024. Most of the variables in the dataset are tokenized to a wide range of [1,7]. We use these statistics to define the $MaxAllowedToken$ parameter in Algorithm \ref{algo1}.

\begin{figure}
\small
  \includegraphics[width=7 cm]{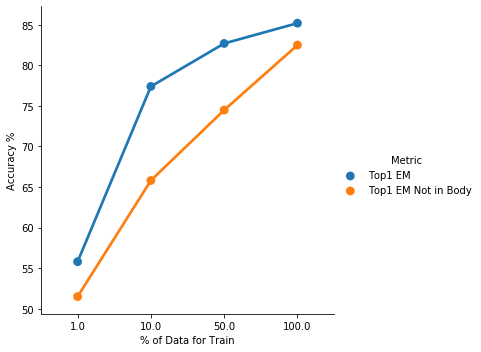}
  \caption{Impact of training corpus size on the performance of our model. The scores are for VarBERT-Small trained with CMLM, 50K Vocab, and Heuristics.}
  \Description{Impact of training corpus size on the performance of our model. The scores are for VarBERT-Small trained with CMLM, 50K Vocab, and Heuristics. Higher is better.}
  \label{fig:learningcurve}
\end{figure}

\begin{figure}
\small
  \includegraphics[width=7 cm]{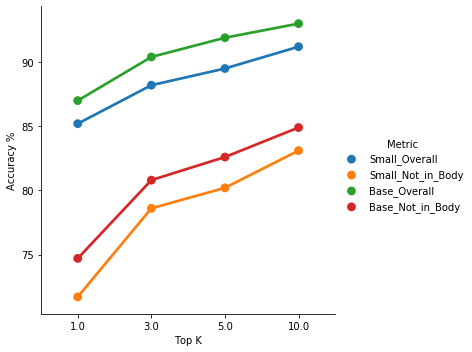}
  \caption{Impact of training corpus size on the performance of our model. The scores are for VarBERT-Small trained with CMLM, 50K Vocab, and Heuristics based Count of Token Prediction.}
  \Description{Accuracy of our VarBERT models trained with CMLM, 50K Vocab and Heuristics based Count of Token Prediction, at different Top K Ranks. }
  \label{fig:topk}
\end{figure}

\begin{figure}
\small
  \includegraphics[width=7 cm]{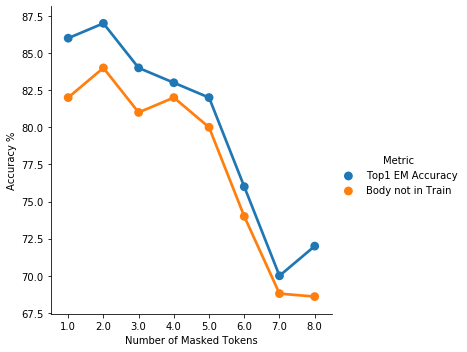}
  \caption{Accuracy of VarBert-Small for different lengths of target variable names. }
  \label{fig:toklenacc}
\end{figure}

\subsection{Model Analysis}
\textbf{Can we use just Masked Language Modeling for Variable Name Recovery?}
In Table \ref{tab:mlmcmlm}, we compare the different modeling tasks. We observe that Constrained Masked Language Modeling with Whole-word Masking performs the best. It is also observed that just Masked Language Modeling performs significantly poor. This can be explained by the random 15\% of tokens which are masked, and not all the variable name tokens. Masked Language Modeling with Whole-word Masking is slightly better. Pre-training also has a significant impact, as without Pre-training model Top1 Exact Match Accuracy are in the lower 50s, with a high perplexity. This is because the task of Constrained Masked Language Modeling is insufficient to learn rich vector representations for the entire vocabulary set.  

\textbf{Does increasing vocabulary size help?}
In Table \ref{tab:vocabcomp} we compare our models with two different sets of vocabulary learned using Byte-Pair Encoding. We observe increasing vocabulary has a significant impact. 
With a smaller vocabulary, we can have a smaller model that trains faster, but we lose in performance by a significant margin.

\begin{table}[]
\small
\begin{tabular}{|l|l|l|l|l|l|}
\hline
Metric & Model & $MLM$ & $MLM_{w}$ & $CMLM$ & $CMLM_{w}$ \\ \hline
\multirow{2}{*}{Top1 EM \% $\uparrow$} & Small & 0 & 1.2 & 83.5 &  85.2 \\ \cline{2-6} 
 & Base & 5 & 5.5 & 86.5 & \textbf{87.0 }\\ \hline
\multirow{2}{*}{Perplexity $\downarrow$} & Small & 1178 & 1033 & 1.07  & \textbf{1.067 }\\ \cline{2-6} 
 & Base & 932 & 899 & 1.58 & 1.081 \\ \hline
\multirow{2}{*}{CER \% $\downarrow$} & Small & 98 & 97.8 & 18.32 &  16.4\\ \cline{2-6} 
 & Base & 97.2 & 96.5 & 16.54 & \textbf{15.65} \\ \hline
\end{tabular}
\caption{Comparison of Masked Language Modeling (MLM) and Constrained MLM (CMLM) on downstream Variable Name Recovery Validation Set. Vocab Size is 50k, and Token Prediction is through heuristics. W refers to being trained with whole-word masking. }
\label{tab:mlmcmlm}
\end{table}

\begin{table}[]
\small
\begin{tabular}{|c|c|c|c|}
\hline
Metric & Model Type & Vocab-20k & Vocab-50k \\ \hline
\multirow{2}{*}{Top1 EM $\uparrow$ } & Small & 78.4 & \textbf{85.2} \\ \cline{2-4} 
 & Base & 80.2 &  \textbf{87.0} \\ \hline
\multirow{2}{*}{Perplexity $\downarrow$ } & Small & 1.832 & \textbf{1.067 }\\ \cline{2-4} 
 & Base & 1.778 & 1.081 \\ \hline
\multirow{2}{*}{CER $\downarrow$} & Small & 19.6 & \textbf{16.4 }\\ \cline{2-4} 
 & Base & 18.78 & \textbf{15.65}\\ \hline
\end{tabular}
\caption{Comparison of Vocabulary Size on downstream Variable Name Recovery Validation Set. Models are finetuned using CMLM, Token Prediction is through heuristics.}
\label{tab:vocabcomp}
\end{table}

\begin{table}[]
\small
\begin{tabular}{|c|c|c|c|}
\hline
Metric & Model Type & Oracle & Heuristics \\ \hline
\multirow{2}{*}{Top1 EM $\uparrow$ } & Small & 91.1 &  85.2\\ \cline{2-4} 
 & Base & 92.6 & 87.0 \\ \hline
\multirow{2}{*}{Perplexity $\downarrow$   } & Small & 1.054 & 1.067 \\ \cline{2-4} 
 & Base & 1.077 & 1.081 \\ \hline
\multirow{2}{*}{CER $\downarrow$} & Small & 11.2 & 16.4 \\ \cline{2-4} 
 & Base & 10.6 & 15.65 \\ \hline
\end{tabular}
\caption{Comparison of Count of Token Prediction algorithm on Variable Name Recovery Validation Set. }
\label{tab:algoimp}
\end{table}

\begin{table}[]
\small
\begin{tabular}{|c|c|c|c|}
\hline
Data Splits & Model Type & Top1 EM $\uparrow$ & CER  $\downarrow$ \\ \hline
\multirow{2}{*}{Overall} & Small &  85.2 &  16.50  \\ \cline{2-4} 
 & Base & 87.0 &  15.65  \\ \hline
\multirow{2}{*}{Body in Train} & Small & 86.7 & 15.40  \\ \cline{2-4} 
 & Base & 89.9 &  14.34  \\ \hline
\multirow{2}{*}{Body not in Train} & Small & 71.8 & 26.70  \\ \cline{2-4} 
 & Base & 74.7 & 24.20 \\ \hline
\end{tabular}
\caption{Comparison of the two trained models, VarBERT-Small and VarBERT-Base on the different splits of the Validation Sets. Models are trained with CMLM and Token prediction heuristics is used to predict count.}
\label{tab:modelbodycomp}
\end{table}

\begin{table}[]
\small
\begin{tabular}{|l|l|l|l|}
\hline
Metric  1\% Data & DEBIN & DIRE & Small \\ \hline
Train Time (h) $\downarrow$& 2.4 & 13 & 3 \\ \hline
Accuracy \% - Overall $\uparrow$ & 2.4 & 32.2  & 55.8 \\ \hline
Accuracy \% - Body in Train  $\uparrow$& 3.0 & 40.0 & 56.8 \\ \hline
Accuracy \% - Body not in Train $\uparrow$ & 0.6 & 5.3 & 51.5 \\ \hline
\end{tabular}
\caption{Comparisons with Baselines Trained on 1\% data. VarBERT-Small is pre-trained with MLM and finetuned with CMLM, with a vocab of 50k and Token Prediction is with heuristics. The scores are on the Validation Set.}
\label{tab:baselines}
\end{table}

\begin{table}[]
\small
\begin{tabular}{|l|l|l|l|}
\hline
Method & Top1 EM \% $\uparrow$ & CER \% $\downarrow$ & Body Not in Train \% $\uparrow$ \\ \hline
DIRE & 74.0 & 28.3 & 35.30 \\ \hline
VarBERT-Small & 83.10 & 17.8 & 82.47 \\ \hline
VarBERT-Base & \textbf{86.36} & \textbf{15.4} & \textbf{84.15} \\ \hline
\end{tabular}
\caption{Comparison with the current state-of-the-art on the Test Set. Body Not In Train is also EM Accuracy.}
\label{tab:sota}
\end{table}

\textbf{How do VarBERT-base and VarBERT-small compared to each other?} Our hypothesis of a smaller model with lesser parameters being able to perform reasonably well is demonstrated across all Tables. The VarBERT-base model does perform better but given a restricted hyper-parameter budget, we observe the delta between the performances is within 2-3 \% in Exact Match Accuracy. It should be noted VarBERT-small has half the number of layers compared to VarBERT-base but has double the max sequence length allowed. This makes VarBERT-small more useful as it can take input longer sequences of code. 

\textbf{How does our Heuristic Based Token Count Prediction Algorithm perform?}
From Table \ref{tab:algoimp} it can be seen that if we know the correct number of masked tokens, the model can predict the original variable name with very high accuracy. The heuristics-based algorithm although performs reasonably well, it has a considerable room to improve. 

\textbf{How does our performance improve if we suggest users with Top K Variable Names?} Figure \ref{fig:topk} shows our performance improves by nearly 7-9\% when we suggest Top K variable names. This shows the promise of using suggestion and ranking based models for variable name recovery.

\textbf{How does our model improve with increasing train data?}
Figure \ref{fig:learningcurve} shows the learning curve of our VarBERT-Small model trained with Constrained Masked Language Modeling and a vocabulary size of 50K. We observe an interesting curve where the overall Top1 accuracy increases more than the \textit{Body not in Train} set. This indicates with more data model learns to memorize the different variable names in shared library functions.

\textbf{How does our models compare to existing baseline and state-of-the-art models?}
Table \ref{tab:baselines} and Table \ref{tab:sota} show our model performs significantly well compared to both the models across all metrics. In Table \ref{tab:baselines}, we compare the baselines with models trained on 1\% of the Training data corpus. We choose this approach as DEBIN requires a considerable amount of training time and does not scale to the entire dataset. We compare with DIRE by training with the entire dataset in Table \ref{tab:sota}. VarBERT-small is more robust and data-efficient compared to both DEBIN and DIRE. Our pre-trained on Masked LM and finetuned on the Constrained MLM model generalizes significantly well, which is shown on the Exact Match Accuracy in the \textit{Body not in Train} set. VarBERT-small is a more accurate and generalizable technique compared to the current state-of-the-art.

\textbf{What is our performance for variables which are split into multiple tokens?}
Figure \ref{fig:toklenacc} shows the accuracy of our model across variables that are split into multiple tokens. It is expected for the model to perform well on variables that are not split as they have rich representations learned during the pre-training task. It is interesting to note that the model performs reasonably well for variables split into two to five tokens. This shows the model has sufficient reasoning capabilities to predict such variable names.

\textbf{Where does our model go wrong?}

We analyzed the different errors the model make and we classify the errors broadly into the following categories: Error in the number of mask count prediction, Partial incorrect token prediction, and Off-by-few-chars errors. Error in the number of mask count prediction occurs when a different set of mask tokens count has a higher average probability. It is observed that a smaller number of tokens have a higher average probability. We can fix this issue if we learn this task instead of a heuristic-based algorithm.

Partial incorrect token prediction happens when a part of the tokens generated is wrong. This happens when there can be several generated combinations with the same prefix token. For example, variable names like \textit{tokenIndex} and \textit{tokenCount}. Our model predicts \textit{tokenIndex} when instead, the original variable is \textit{tokenCount}. This can be corrected by a suggestion based system and it gives the user the freedom to select which variable name is more suited.

Off-by-few-chars errors occur when the model predicts tokens which only differ by a few characters. For example, \textit{token} and \textit{tokens}. Both are present in our vocabulary, and the model predicts both in the top two positions together. There are several such pairs, and not restricted to only a single length of split-tokens.

\textbf{What are the other sources of errors?}
The DIRE dataset comprises of scraped C code from GitHub. There was no filtering and quality control implemented to ensure that the dataset represents the set of target binaries that are reverse-engineered. 
Moreover, these binaries were not compiled with multiple different optimization and obfuscation compiler options. There is a possibility that our models may not perform on such code with such high accuracy but may work reasonably well compared to our 
baselines, and current state-of-the-art. To improve on such binaries is left as future work.
Our model is tightly coupled with the output of the Hex-Rays decompiler. To support other decompilers, the model may need to be re-trained with a new training corpus generated from the new set of decompilers. We leave building an adaptable and multi-decompiler supporting model as future work.  

\section{Conclusion}

Improving the understandability of decompiled binary code is crucial for software reverse engineering.
In this paper, we have advanced the state of the art for variable name recovery in decompiled binary code.
We adapted recent advances in the field of natural language processing, such as Masked Language Modeling and Transformers, and proposed a heuristic-based Count of Token Prediction algorithm.
These techniques are crucial in improving the performance of variable name recovery in decompiled binary code.

In our evaluation, we showed the impact of each module.
We trained two neural network models, VarBERT-base and VarBERT-small.
These neural models take \textit{raw} code as input, which makes them much simpler to build and use.
Our evaluation of the DIRE data set shows that our techniques advance the state-of-the-art by 12.36\% on overall accuracy and improve on generalizability by 49\%.


\bibliographystyle{ACM-Reference-Format}
\bibliography{sample-base}


\begin{thebibliography}{54}


\ifx \showCODEN    \undefined \def \showCODEN     #1{\unskip}     \fi
\ifx \showDOI      \undefined \def \showDOI       #1{#1}\fi
\ifx \showISBNx    \undefined \def \showISBNx     #1{\unskip}     \fi
\ifx \showISBNxiii \undefined \def \showISBNxiii  #1{\unskip}     \fi
\ifx \showISSN     \undefined \def \showISSN      #1{\unskip}     \fi
\ifx \showLCCN     \undefined \def \showLCCN      #1{\unskip}     \fi
\ifx \shownote     \undefined \def \shownote      #1{#1}          \fi
\ifx \showarticletitle \undefined \def \showarticletitle #1{#1}   \fi
\ifx \showURL      \undefined \def \showURL       {\relax}        \fi
\providecommand\bibfield[2]{#2}
\providecommand\bibinfo[2]{#2}
\providecommand\natexlab[1]{#1}
\providecommand\showeprint[2][]{arXiv:#2}

\bibitem[\protect\citeauthoryear{Allamanis, Barr, Bird, and Sutton}{Allamanis
  et~al\mbox{.}}{2014}]%
        {allamanis2014learning}
\bibfield{author}{\bibinfo{person}{Miltiadis Allamanis},
  \bibinfo{person}{Earl~T Barr}, \bibinfo{person}{Christian Bird}, {and}
  \bibinfo{person}{Charles Sutton}.} \bibinfo{year}{2014}\natexlab{}.
\newblock \showarticletitle{Learning natural coding conventions}. In
  \bibinfo{booktitle}{\emph{Proceedings of the 22nd ACM SIGSOFT International
  Symposium on Foundations of Software Engineering}}.
  \bibinfo{pages}{281--293}.
\newblock


\bibitem[\protect\citeauthoryear{Allamanis, Barr, Devanbu, and
  Sutton}{Allamanis et~al\mbox{.}}{2018}]%
        {allamanis2018survey}
\bibfield{author}{\bibinfo{person}{Miltiadis Allamanis},
  \bibinfo{person}{Earl~T Barr}, \bibinfo{person}{Premkumar Devanbu}, {and}
  \bibinfo{person}{Charles Sutton}.} \bibinfo{year}{2018}\natexlab{}.
\newblock \showarticletitle{A survey of machine learning for big code and
  naturalness}.
\newblock \bibinfo{journal}{\emph{ACM Computing Surveys (CSUR)}}
  \bibinfo{volume}{51}, \bibinfo{number}{4} (\bibinfo{year}{2018}),
  \bibinfo{pages}{1--37}.
\newblock


\bibitem[\protect\citeauthoryear{Artuso, Di~Luna, Massarelli, and
  Querzoni}{Artuso et~al\mbox{.}}{2019}]%
        {artuso2019function}
\bibfield{author}{\bibinfo{person}{Fiorella Artuso},
  \bibinfo{person}{Giuseppe~Antonio Di~Luna}, \bibinfo{person}{Luca
  Massarelli}, {and} \bibinfo{person}{Leonardo Querzoni}.}
  \bibinfo{year}{2019}\natexlab{}.
\newblock \showarticletitle{Function Naming in Stripped Binaries Using Neural
  Networks}.
\newblock \bibinfo{journal}{\emph{arXiv preprint arXiv:1912.07946}}
  (\bibinfo{year}{2019}).
\newblock


\bibitem[\protect\citeauthoryear{Bavishi, Pradel, and Sen}{Bavishi
  et~al\mbox{.}}{2018}]%
        {bavishi2018context2name}
\bibfield{author}{\bibinfo{person}{Rohan Bavishi}, \bibinfo{person}{Michael
  Pradel}, {and} \bibinfo{person}{Koushik Sen}.}
  \bibinfo{year}{2018}\natexlab{}.
\newblock \showarticletitle{Context2Name: A deep learning-based approach to
  infer natural variable names from usage contexts}.
\newblock \bibinfo{journal}{\emph{arXiv preprint arXiv:1809.05193}}
  (\bibinfo{year}{2018}).
\newblock


\bibitem[\protect\citeauthoryear{Beltramelli}{Beltramelli}{2018}]%
        {beltramelli2018pix2code}
\bibfield{author}{\bibinfo{person}{Tony Beltramelli}.}
  \bibinfo{year}{2018}\natexlab{}.
\newblock \showarticletitle{pix2code: Generating code from a graphical user
  interface screenshot}. In \bibinfo{booktitle}{\emph{Proceedings of the ACM
  SIGCHI Symposium on Engineering Interactive Computing Systems}}.
  \bibinfo{pages}{1--6}.
\newblock


\bibitem[\protect\citeauthoryear{Bhatia and Singh}{Bhatia and Singh}{2016}]%
        {bhatia2016automated}
\bibfield{author}{\bibinfo{person}{Sahil Bhatia} {and} \bibinfo{person}{Rishabh
  Singh}.} \bibinfo{year}{2016}\natexlab{}.
\newblock \showarticletitle{Automated correction for syntax errors in
  programming assignments using recurrent neural networks}.
\newblock \bibinfo{journal}{\emph{arXiv preprint arXiv:1603.06129}}
  (\bibinfo{year}{2016}).
\newblock


\bibitem[\protect\citeauthoryear{Bruch, Monperrus, and Mezini}{Bruch
  et~al\mbox{.}}{2009}]%
        {bruch2009learning}
\bibfield{author}{\bibinfo{person}{Marcel Bruch}, \bibinfo{person}{Martin
  Monperrus}, {and} \bibinfo{person}{Mira Mezini}.}
  \bibinfo{year}{2009}\natexlab{}.
\newblock \showarticletitle{Learning from examples to improve code completion
  systems}. In \bibinfo{booktitle}{\emph{Proceedings of the 7th joint meeting
  of the European software engineering conference and the ACM SIGSOFT symposium
  on the foundations of software engineering}}. \bibinfo{pages}{213--222}.
\newblock


\bibitem[\protect\citeauthoryear{Caballero and Lin}{Caballero and Lin}{2016}]%
        {Caballero2016}
\bibfield{author}{\bibinfo{person}{Juan Caballero} {and}
  \bibinfo{person}{Zhiqiang Lin}.} \bibinfo{year}{2016}\natexlab{}.
\newblock \showarticletitle{Type Inference on Executables}.
\newblock \bibinfo{journal}{\emph{Comput. Surveys}} \bibinfo{volume}{48},
  \bibinfo{number}{4} (\bibinfo{year}{2016}), \bibinfo{pages}{1--35}.
\newblock
\showISSN{03600300}
\urldef\tempurl%
\url{https://doi.org/10.1145/2896499}
\showDOI{\tempurl}


\bibitem[\protect\citeauthoryear{Campbell, Hindle, and Amaral}{Campbell
  et~al\mbox{.}}{2014}]%
        {campbell2014syntax}
\bibfield{author}{\bibinfo{person}{Joshua~Charles Campbell},
  \bibinfo{person}{Abram Hindle}, {and} \bibinfo{person}{Jos{\'e}~Nelson
  Amaral}.} \bibinfo{year}{2014}\natexlab{}.
\newblock \showarticletitle{Syntax errors just aren't natural: improving error
  reporting with language models}. In \bibinfo{booktitle}{\emph{Proceedings of
  the 11th Working Conference on Mining Software Repositories}}.
  \bibinfo{pages}{252--261}.
\newblock


\bibitem[\protect\citeauthoryear{Cerulo, Ceccarelli, Di~Penta, and
  Canfora}{Cerulo et~al\mbox{.}}{2013}]%
        {cerulo2013hidden}
\bibfield{author}{\bibinfo{person}{Luigi Cerulo}, \bibinfo{person}{Michele
  Ceccarelli}, \bibinfo{person}{Massimiliano Di~Penta}, {and}
  \bibinfo{person}{Gerardo Canfora}.} \bibinfo{year}{2013}\natexlab{}.
\newblock \showarticletitle{A hidden markov model to detect coded information
  islands in free text}. In \bibinfo{booktitle}{\emph{2013 IEEE 13th
  International Working Conference on Source Code Analysis and Manipulation
  (SCAM)}}. IEEE, \bibinfo{pages}{157--166}.
\newblock


\bibitem[\protect\citeauthoryear{Chagnon}{Chagnon}{[n.d.]}]%
        {chagnonida}
\bibfield{author}{\bibinfo{person}{F Chagnon}.}
  \bibinfo{year}{[n.d.]}\natexlab{}.
\newblock \bibinfo{title}{IDA-Decompiler}.
\newblock
\newblock


\bibitem[\protect\citeauthoryear{David, Alon, and Yahav}{David
  et~al\mbox{.}}{2019}]%
        {david2019neural}
\bibfield{author}{\bibinfo{person}{Yaniv David}, \bibinfo{person}{Uri Alon},
  {and} \bibinfo{person}{Eran Yahav}.} \bibinfo{year}{2019}\natexlab{}.
\newblock \showarticletitle{Neural reverse engineering of stripped binaries}.
\newblock \bibinfo{journal}{\emph{arXiv preprint arXiv:1902.09122}}
  (\bibinfo{year}{2019}).
\newblock


\bibitem[\protect\citeauthoryear{Devanbu}{Devanbu}{2015}]%
        {devanbu2015new}
\bibfield{author}{\bibinfo{person}{Premkumar Devanbu}.}
  \bibinfo{year}{2015}\natexlab{}.
\newblock \showarticletitle{New initiative: the naturalness of software}. In
  \bibinfo{booktitle}{\emph{2015 IEEE/ACM 37th IEEE International Conference on
  Software Engineering}}, Vol.~\bibinfo{volume}{2}. IEEE,
  \bibinfo{pages}{543--546}.
\newblock


\bibitem[\protect\citeauthoryear{Devlin, Chang, Lee, and Toutanova}{Devlin
  et~al\mbox{.}}{2018}]%
        {devlin2018bert}
\bibfield{author}{\bibinfo{person}{Jacob Devlin}, \bibinfo{person}{Ming-Wei
  Chang}, \bibinfo{person}{Kenton Lee}, {and} \bibinfo{person}{Kristina
  Toutanova}.} \bibinfo{year}{2018}\natexlab{}.
\newblock \showarticletitle{Bert: Pre-training of deep bidirectional
  transformers for language understanding}.
\newblock \bibinfo{journal}{\emph{arXiv preprint arXiv:1810.04805}}
  (\bibinfo{year}{2018}).
\newblock


\bibitem[\protect\citeauthoryear{ghidra decompiler}{ghidra decompiler}{2019}]%
        {ghidra}
\bibfield{author}{\bibinfo{person}{The ghidra decompiler}.}
  \bibinfo{year}{2019}\natexlab{}.
\newblock \bibinfo{title}{{The ghidra decompiler}}.
\newblock
\newblock
\urldef\tempurl%
\url{https://ghidra-sre.org/}
\showURL{%
\tempurl}


\bibitem[\protect\citeauthoryear{Gupta, Kanade, and Shevade}{Gupta
  et~al\mbox{.}}{2018}]%
        {gupta2018deep}
\bibfield{author}{\bibinfo{person}{Rahul Gupta}, \bibinfo{person}{Aditya
  Kanade}, {and} \bibinfo{person}{Shirish Shevade}.}
  \bibinfo{year}{2018}\natexlab{}.
\newblock \showarticletitle{Deep reinforcement learning for programming
  language correction}.
\newblock \bibinfo{journal}{\emph{arXiv preprint arXiv:1801.10467}}
  (\bibinfo{year}{2018}).
\newblock


\bibitem[\protect\citeauthoryear{Gupta, Pal, Kanade, and Shevade}{Gupta
  et~al\mbox{.}}{2017}]%
        {gupta2017deepfix}
\bibfield{author}{\bibinfo{person}{Rahul Gupta}, \bibinfo{person}{Soham Pal},
  \bibinfo{person}{Aditya Kanade}, {and} \bibinfo{person}{Shirish Shevade}.}
  \bibinfo{year}{2017}\natexlab{}.
\newblock \showarticletitle{Deepfix: Fixing common c language errors by deep
  learning}. In \bibinfo{booktitle}{\emph{Thirty-First AAAI Conference on
  Artificial Intelligence}}.
\newblock


\bibitem[\protect\citeauthoryear{Gvero and Kuncak}{Gvero and Kuncak}{2015}]%
        {gvero2015synthesizing}
\bibfield{author}{\bibinfo{person}{Tihomir Gvero} {and} \bibinfo{person}{Viktor
  Kuncak}.} \bibinfo{year}{2015}\natexlab{}.
\newblock \showarticletitle{Synthesizing Java expressions from free-form
  queries}. In \bibinfo{booktitle}{\emph{Proceedings of the 2015 ACM SIGPLAN
  International Conference on Object-Oriented Programming, Systems, Languages,
  and Applications}}. \bibinfo{pages}{416--432}.
\newblock


\bibitem[\protect\citeauthoryear{He, Ivanov, Tsankov, Raychev, and Vechev}{He
  et~al\mbox{.}}{2018}]%
        {he2018debin}
\bibfield{author}{\bibinfo{person}{Jingxuan He}, \bibinfo{person}{Pesho
  Ivanov}, \bibinfo{person}{Petar Tsankov}, \bibinfo{person}{Veselin Raychev},
  {and} \bibinfo{person}{Martin Vechev}.} \bibinfo{year}{2018}\natexlab{}.
\newblock \showarticletitle{Debin: Predicting debug information in stripped
  binaries}. In \bibinfo{booktitle}{\emph{Proceedings of the 2018 ACM SIGSAC
  Conference on Computer and Communications Security}}.
  \bibinfo{pages}{1667--1680}.
\newblock


\bibitem[\protect\citeauthoryear{Hendrycks and Gimpel}{Hendrycks and
  Gimpel}{2016}]%
        {hendrycks2016gaussian}
\bibfield{author}{\bibinfo{person}{Dan Hendrycks} {and} \bibinfo{person}{Kevin
  Gimpel}.} \bibinfo{year}{2016}\natexlab{}.
\newblock \showarticletitle{Gaussian error linear units (gelus)}.
\newblock \bibinfo{journal}{\emph{arXiv preprint arXiv:1606.08415}}
  (\bibinfo{year}{2016}).
\newblock


\bibitem[\protect\citeauthoryear{Hex-Rays}{Hex-Rays}{2013}]%
        {hex2013hex}
\bibfield{author}{\bibinfo{person}{SA Hex-Rays}.}
  \bibinfo{year}{2013}\natexlab{}.
\newblock \bibinfo{title}{Hex-Rays Decompiler}.
\newblock
\newblock


\bibitem[\protect\citeauthoryear{Hindle, Barr, Su, Gabel, and Devanbu}{Hindle
  et~al\mbox{.}}{2012}]%
        {hindle2012naturalness}
\bibfield{author}{\bibinfo{person}{Abram Hindle}, \bibinfo{person}{Earl~T
  Barr}, \bibinfo{person}{Zhendong Su}, \bibinfo{person}{Mark Gabel}, {and}
  \bibinfo{person}{Premkumar Devanbu}.} \bibinfo{year}{2012}\natexlab{}.
\newblock \showarticletitle{On the naturalness of software}. In
  \bibinfo{booktitle}{\emph{2012 34th International Conference on Software
  Engineering (ICSE)}}. IEEE, \bibinfo{pages}{837--847}.
\newblock


\bibitem[\protect\citeauthoryear{Hochreiter and Schmidhuber}{Hochreiter and
  Schmidhuber}{1997}]%
        {hochreiter1997long}
\bibfield{author}{\bibinfo{person}{Sepp Hochreiter} {and}
  \bibinfo{person}{J{\"u}rgen Schmidhuber}.} \bibinfo{year}{1997}\natexlab{}.
\newblock \showarticletitle{Long short-term memory}.
\newblock \bibinfo{journal}{\emph{Neural computation}} \bibinfo{volume}{9},
  \bibinfo{number}{8} (\bibinfo{year}{1997}), \bibinfo{pages}{1735--1780}.
\newblock


\bibitem[\protect\citeauthoryear{Hu, Wei, Li, and Jin}{Hu
  et~al\mbox{.}}{2017}]%
        {hu2017codesum}
\bibfield{author}{\bibinfo{person}{Xing Hu}, \bibinfo{person}{Yuhan Wei},
  \bibinfo{person}{Ge Li}, {and} \bibinfo{person}{Zhi Jin}.}
  \bibinfo{year}{2017}\natexlab{}.
\newblock \showarticletitle{CodeSum: Translate program language to natural
  language}.
\newblock \bibinfo{journal}{\emph{arXiv preprint arXiv:1708.01837}}
  (\bibinfo{year}{2017}).
\newblock


\bibitem[\protect\citeauthoryear{Iyer, Konstas, Cheung, and Zettlemoyer}{Iyer
  et~al\mbox{.}}{2016}]%
        {iyer2016summarizing}
\bibfield{author}{\bibinfo{person}{Srinivasan Iyer}, \bibinfo{person}{Ioannis
  Konstas}, \bibinfo{person}{Alvin Cheung}, {and} \bibinfo{person}{Luke
  Zettlemoyer}.} \bibinfo{year}{2016}\natexlab{}.
\newblock \showarticletitle{Summarizing source code using a neural attention
  model}. In \bibinfo{booktitle}{\emph{Proceedings of the 54th Annual Meeting
  of the Association for Computational Linguistics (Volume 1: Long Papers)}}.
  \bibinfo{pages}{2073--2083}.
\newblock


\bibitem[\protect\citeauthoryear{Jaffe, Lacomis, Schwartz, Goues, and
  Vasilescu}{Jaffe et~al\mbox{.}}{2018}]%
        {jaffe2018meaningful}
\bibfield{author}{\bibinfo{person}{Alan Jaffe}, \bibinfo{person}{Jeremy
  Lacomis}, \bibinfo{person}{Edward~J Schwartz}, \bibinfo{person}{Claire~Le
  Goues}, {and} \bibinfo{person}{Bogdan Vasilescu}.}
  \bibinfo{year}{2018}\natexlab{}.
\newblock \showarticletitle{Meaningful variable names for decompiled code: A
  machine translation approach}. In \bibinfo{booktitle}{\emph{Proceedings of
  the 26th Conference on Program Comprehension}}. \bibinfo{pages}{20--30}.
\newblock


\bibitem[\protect\citeauthoryear{Kingma and Ba}{Kingma and Ba}{2014}]%
        {kingma2014adam}
\bibfield{author}{\bibinfo{person}{Diederik~P Kingma} {and}
  \bibinfo{person}{Jimmy Ba}.} \bibinfo{year}{2014}\natexlab{}.
\newblock \showarticletitle{Adam: A method for stochastic optimization}.
\newblock \bibinfo{journal}{\emph{arXiv preprint arXiv:1412.6980}}
  (\bibinfo{year}{2014}).
\newblock


\bibitem[\protect\citeauthoryear{Kudo and Richardson}{Kudo and
  Richardson}{2018}]%
        {kudo2018sentencepiece}
\bibfield{author}{\bibinfo{person}{Taku Kudo} {and} \bibinfo{person}{John
  Richardson}.} \bibinfo{year}{2018}\natexlab{}.
\newblock \showarticletitle{Sentencepiece: A simple and language independent
  subword tokenizer and detokenizer for neural text processing}.
\newblock \bibinfo{journal}{\emph{arXiv preprint arXiv:1808.06226}}
  (\bibinfo{year}{2018}).
\newblock


\bibitem[\protect\citeauthoryear{Kushman and Barzilay}{Kushman and
  Barzilay}{2013}]%
        {kushman2013using}
\bibfield{author}{\bibinfo{person}{Nate Kushman} {and} \bibinfo{person}{Regina
  Barzilay}.} \bibinfo{year}{2013}\natexlab{}.
\newblock \showarticletitle{Using semantic unification to generate regular
  expressions from natural language}. In \bibinfo{booktitle}{\emph{Proceedings
  of the 2013 Conference of the North American Chapter of the Association for
  Computational Linguistics: Human Language Technologies}}.
  \bibinfo{pages}{826--836}.
\newblock


\bibitem[\protect\citeauthoryear{Lacomis, Yin, Schwartz, Allamanis, Le~Goues,
  Neubig, and Vasilescu}{Lacomis et~al\mbox{.}}{2019}]%
        {lacomis2019dire}
\bibfield{author}{\bibinfo{person}{Jeremy Lacomis}, \bibinfo{person}{Pengcheng
  Yin}, \bibinfo{person}{Edward Schwartz}, \bibinfo{person}{Miltiadis
  Allamanis}, \bibinfo{person}{Claire Le~Goues}, \bibinfo{person}{Graham
  Neubig}, {and} \bibinfo{person}{Bogdan Vasilescu}.}
  \bibinfo{year}{2019}\natexlab{}.
\newblock \showarticletitle{Dire: A neural approach to decompiled identifier
  naming}. In \bibinfo{booktitle}{\emph{2019 34th IEEE/ACM International
  Conference on Automated Software Engineering (ASE)}}. IEEE,
  \bibinfo{pages}{628--639}.
\newblock


\bibitem[\protect\citeauthoryear{Li, Tarlow, Brockschmidt, and Zemel}{Li
  et~al\mbox{.}}{2015}]%
        {li2015gated}
\bibfield{author}{\bibinfo{person}{Yujia Li}, \bibinfo{person}{Daniel Tarlow},
  \bibinfo{person}{Marc Brockschmidt}, {and} \bibinfo{person}{Richard Zemel}.}
  \bibinfo{year}{2015}\natexlab{}.
\newblock \showarticletitle{Gated graph sequence neural networks}.
\newblock \bibinfo{journal}{\emph{arXiv preprint arXiv:1511.05493}}
  (\bibinfo{year}{2015}).
\newblock


\bibitem[\protect\citeauthoryear{Ling, Grefenstette, Hermann,
  Ko{\v{c}}isk{\`y}, Senior, Wang, and Blunsom}{Ling et~al\mbox{.}}{2016}]%
        {ling2016latent}
\bibfield{author}{\bibinfo{person}{Wang Ling}, \bibinfo{person}{Edward
  Grefenstette}, \bibinfo{person}{Karl~Moritz Hermann},
  \bibinfo{person}{Tom{\'a}{\v{s}} Ko{\v{c}}isk{\`y}}, \bibinfo{person}{Andrew
  Senior}, \bibinfo{person}{Fumin Wang}, {and} \bibinfo{person}{Phil Blunsom}.}
  \bibinfo{year}{2016}\natexlab{}.
\newblock \showarticletitle{Latent predictor networks for code generation}.
\newblock \bibinfo{journal}{\emph{arXiv preprint arXiv:1603.06744}}
  (\bibinfo{year}{2016}).
\newblock


\bibitem[\protect\citeauthoryear{Liu}{Liu}{2016}]%
        {liu2016towards}
\bibfield{author}{\bibinfo{person}{Han Liu}.} \bibinfo{year}{2016}\natexlab{}.
\newblock \showarticletitle{Towards better program obfuscation: optimization
  via language models}. In \bibinfo{booktitle}{\emph{2016 IEEE/ACM 38th
  International Conference on Software Engineering Companion (ICSE-C)}}. IEEE,
  \bibinfo{pages}{680--682}.
\newblock


\bibitem[\protect\citeauthoryear{Liu, Ott, Goyal, Du, Joshi, Chen, Levy, Lewis,
  Zettlemoyer, and Stoyanov}{Liu et~al\mbox{.}}{2019}]%
        {liu2019roberta}
\bibfield{author}{\bibinfo{person}{Yinhan Liu}, \bibinfo{person}{Myle Ott},
  \bibinfo{person}{Naman Goyal}, \bibinfo{person}{Jingfei Du},
  \bibinfo{person}{Mandar Joshi}, \bibinfo{person}{Danqi Chen},
  \bibinfo{person}{Omer Levy}, \bibinfo{person}{Mike Lewis},
  \bibinfo{person}{Luke Zettlemoyer}, {and} \bibinfo{person}{Veselin
  Stoyanov}.} \bibinfo{year}{2019}\natexlab{}.
\newblock \showarticletitle{Roberta: A robustly optimized bert pretraining
  approach}.
\newblock \bibinfo{journal}{\emph{arXiv preprint arXiv:1907.11692}}
  (\bibinfo{year}{2019}).
\newblock


\bibitem[\protect\citeauthoryear{Omar}{Omar}{2013}]%
        {omar2013structured}
\bibfield{author}{\bibinfo{person}{Cyrus Omar}.}
  \bibinfo{year}{2013}\natexlab{}.
\newblock \showarticletitle{Structured statistical syntax tree prediction}. In
  \bibinfo{booktitle}{\emph{Proceedings of the 2013 companion publication for
  conference on Systems, programming, \& applications: software for humanity}}.
  \bibinfo{pages}{113--114}.
\newblock


\bibitem[\protect\citeauthoryear{Proksch, Lerch, and Mezini}{Proksch
  et~al\mbox{.}}{2015}]%
        {proksch2015intelligent}
\bibfield{author}{\bibinfo{person}{Sebastian Proksch},
  \bibinfo{person}{Johannes Lerch}, {and} \bibinfo{person}{Mira Mezini}.}
  \bibinfo{year}{2015}\natexlab{}.
\newblock \showarticletitle{Intelligent code completion with Bayesian
  networks}.
\newblock \bibinfo{journal}{\emph{ACM Transactions on Software Engineering and
  Methodology (TOSEM)}} \bibinfo{volume}{25}, \bibinfo{number}{1}
  (\bibinfo{year}{2015}), \bibinfo{pages}{1--31}.
\newblock


\bibitem[\protect\citeauthoryear{Pu, Narasimhan, Solar-Lezama, and Barzilay}{Pu
  et~al\mbox{.}}{2016}]%
        {pu2016sk_p}
\bibfield{author}{\bibinfo{person}{Yewen Pu}, \bibinfo{person}{Karthik
  Narasimhan}, \bibinfo{person}{Armando Solar-Lezama}, {and}
  \bibinfo{person}{Regina Barzilay}.} \bibinfo{year}{2016}\natexlab{}.
\newblock \showarticletitle{sk\_p: a neural program corrector for MOOCs}. In
  \bibinfo{booktitle}{\emph{Companion Proceedings of the 2016 ACM SIGPLAN
  International Conference on Systems, Programming, Languages and Applications:
  Software for Humanity}}. \bibinfo{pages}{39--40}.
\newblock


\bibitem[\protect\citeauthoryear{Rabinovich, Stern, and Klein}{Rabinovich
  et~al\mbox{.}}{2017}]%
        {rabinovich2017abstract}
\bibfield{author}{\bibinfo{person}{Maxim Rabinovich}, \bibinfo{person}{Mitchell
  Stern}, {and} \bibinfo{person}{Dan Klein}.} \bibinfo{year}{2017}\natexlab{}.
\newblock \showarticletitle{Abstract syntax networks for code generation and
  semantic parsing}.
\newblock \bibinfo{journal}{\emph{arXiv preprint arXiv:1704.07535}}
  (\bibinfo{year}{2017}).
\newblock


\bibitem[\protect\citeauthoryear{Radford, Wu, Child, Luan, Amodei, and
  Sutskever}{Radford et~al\mbox{.}}{2019}]%
        {radford2019language}
\bibfield{author}{\bibinfo{person}{Alec Radford}, \bibinfo{person}{Jeffrey Wu},
  \bibinfo{person}{Rewon Child}, \bibinfo{person}{David Luan},
  \bibinfo{person}{Dario Amodei}, {and} \bibinfo{person}{Ilya Sutskever}.}
  \bibinfo{year}{2019}\natexlab{}.
\newblock \showarticletitle{Language models are unsupervised multitask
  learners}.
\newblock \bibinfo{journal}{\emph{OpenAI Blog}} \bibinfo{volume}{1},
  \bibinfo{number}{8} (\bibinfo{year}{2019}), \bibinfo{pages}{9}.
\newblock


\bibitem[\protect\citeauthoryear{Ray, Hellendoorn, Godhane, Tu, Bacchelli, and
  Devanbu}{Ray et~al\mbox{.}}{2016}]%
        {ray2016naturalness}
\bibfield{author}{\bibinfo{person}{Baishakhi Ray}, \bibinfo{person}{Vincent
  Hellendoorn}, \bibinfo{person}{Saheel Godhane}, \bibinfo{person}{Zhaopeng
  Tu}, \bibinfo{person}{Alberto Bacchelli}, {and} \bibinfo{person}{Premkumar
  Devanbu}.} \bibinfo{year}{2016}\natexlab{}.
\newblock \showarticletitle{On the" naturalness" of buggy code}. In
  \bibinfo{booktitle}{\emph{2016 IEEE/ACM 38th International Conference on
  Software Engineering (ICSE)}}. IEEE, \bibinfo{pages}{428--439}.
\newblock


\bibitem[\protect\citeauthoryear{Raychev, Vechev, and Krause}{Raychev
  et~al\mbox{.}}{2015}]%
        {raychev2015predicting}
\bibfield{author}{\bibinfo{person}{Veselin Raychev}, \bibinfo{person}{Martin
  Vechev}, {and} \bibinfo{person}{Andreas Krause}.}
  \bibinfo{year}{2015}\natexlab{}.
\newblock \showarticletitle{Predicting program properties from" big code"}.
\newblock \bibinfo{journal}{\emph{ACM SIGPLAN Notices}} \bibinfo{volume}{50},
  \bibinfo{number}{1} (\bibinfo{year}{2015}), \bibinfo{pages}{111--124}.
\newblock


\bibitem[\protect\citeauthoryear{Sennrich, Haddow, and Birch}{Sennrich
  et~al\mbox{.}}{2015}]%
        {sennrich2015neural}
\bibfield{author}{\bibinfo{person}{Rico Sennrich}, \bibinfo{person}{Barry
  Haddow}, {and} \bibinfo{person}{Alexandra Birch}.}
  \bibinfo{year}{2015}\natexlab{}.
\newblock \showarticletitle{Neural machine translation of rare words with
  subword units}.
\newblock \bibinfo{journal}{\emph{arXiv preprint arXiv:1508.07909}}
  (\bibinfo{year}{2015}).
\newblock


\bibitem[\protect\citeauthoryear{Sharma, Tian, and Lo}{Sharma
  et~al\mbox{.}}{2015}]%
        {sharma2015nirmal}
\bibfield{author}{\bibinfo{person}{Abhishek Sharma}, \bibinfo{person}{Yuan
  Tian}, {and} \bibinfo{person}{David Lo}.} \bibinfo{year}{2015}\natexlab{}.
\newblock \showarticletitle{NIRMAL: Automatic identification of software
  relevant tweets leveraging language model}. In \bibinfo{booktitle}{\emph{2015
  IEEE 22nd International Conference on Software Analysis, Evolution, and
  Reengineering (SANER)}}. IEEE, \bibinfo{pages}{449--458}.
\newblock


\bibitem[\protect\citeauthoryear{Tan and Bansal}{Tan and Bansal}{2019}]%
        {tan2019lxmert}
\bibfield{author}{\bibinfo{person}{Hao Tan} {and} \bibinfo{person}{Mohit
  Bansal}.} \bibinfo{year}{2019}\natexlab{}.
\newblock \showarticletitle{Lxmert: Learning cross-modality encoder
  representations from transformers}.
\newblock \bibinfo{journal}{\emph{arXiv preprint arXiv:1908.07490}}
  (\bibinfo{year}{2019}).
\newblock


\bibitem[\protect\citeauthoryear{Vasilescu, Casalnuovo, and Devanbu}{Vasilescu
  et~al\mbox{.}}{2017}]%
        {vasilescu2017recovering}
\bibfield{author}{\bibinfo{person}{Bogdan Vasilescu}, \bibinfo{person}{Casey
  Casalnuovo}, {and} \bibinfo{person}{Premkumar Devanbu}.}
  \bibinfo{year}{2017}\natexlab{}.
\newblock \showarticletitle{Recovering clear, natural identifiers from
  obfuscated JS names}. In \bibinfo{booktitle}{\emph{Proceedings of the 2017
  11th Joint Meeting on Foundations of Software Engineering}}.
  \bibinfo{pages}{683--693}.
\newblock


\bibitem[\protect\citeauthoryear{Vaswani, Shazeer, Parmar, Uszkoreit, Jones,
  Gomez, Kaiser, and Polosukhin}{Vaswani et~al\mbox{.}}{2017}]%
        {vaswani2017attention}
\bibfield{author}{\bibinfo{person}{Ashish Vaswani}, \bibinfo{person}{Noam
  Shazeer}, \bibinfo{person}{Niki Parmar}, \bibinfo{person}{Jakob Uszkoreit},
  \bibinfo{person}{Llion Jones}, \bibinfo{person}{Aidan~N Gomez},
  \bibinfo{person}{{\L}ukasz Kaiser}, {and} \bibinfo{person}{Illia
  Polosukhin}.} \bibinfo{year}{2017}\natexlab{}.
\newblock \showarticletitle{Attention is all you need}. In
  \bibinfo{booktitle}{\emph{Advances in neural information processing
  systems}}. \bibinfo{pages}{5998--6008}.
\newblock


\bibitem[\protect\citeauthoryear{Wang, Chollak, Movshovitz-Attias, and
  Tan}{Wang et~al\mbox{.}}{2016a}]%
        {wang2016bugram}
\bibfield{author}{\bibinfo{person}{Song Wang}, \bibinfo{person}{Devin Chollak},
  \bibinfo{person}{Dana Movshovitz-Attias}, {and} \bibinfo{person}{Lin Tan}.}
  \bibinfo{year}{2016}\natexlab{a}.
\newblock \showarticletitle{Bugram: bug detection with n-gram language models}.
  In \bibinfo{booktitle}{\emph{Proceedings of the 31st IEEE/ACM International
  Conference on Automated Software Engineering}}. \bibinfo{pages}{708--719}.
\newblock


\bibitem[\protect\citeauthoryear{Wang, Peter, Rosendahl, and Ney}{Wang
  et~al\mbox{.}}{2016b}]%
        {wang2016character}
\bibfield{author}{\bibinfo{person}{Weiyue Wang}, \bibinfo{person}{Jan-Thorsten
  Peter}, \bibinfo{person}{Hendrik Rosendahl}, {and} \bibinfo{person}{Hermann
  Ney}.} \bibinfo{year}{2016}\natexlab{b}.
\newblock \showarticletitle{Character: Translation edit rate on character
  level}. In \bibinfo{booktitle}{\emph{Proceedings of the First Conference on
  Machine Translation: Volume 2, Shared Task Papers}}.
  \bibinfo{pages}{505--510}.
\newblock


\bibitem[\protect\citeauthoryear{White, Tufano, Vendome, and Poshyvanyk}{White
  et~al\mbox{.}}{2016}]%
        {white2016deep}
\bibfield{author}{\bibinfo{person}{Martin White}, \bibinfo{person}{Michele
  Tufano}, \bibinfo{person}{Christopher Vendome}, {and} \bibinfo{person}{Denys
  Poshyvanyk}.} \bibinfo{year}{2016}\natexlab{}.
\newblock \showarticletitle{Deep learning code fragments for code clone
  detection}. In \bibinfo{booktitle}{\emph{2016 31st IEEE/ACM International
  Conference on Automated Software Engineering (ASE)}}. IEEE,
  \bibinfo{pages}{87--98}.
\newblock


\bibitem[\protect\citeauthoryear{Wu, Schuster, Chen, Le, Norouzi, Macherey,
  Krikun, Cao, Gao, Macherey, et~al\mbox{.}}{Wu et~al\mbox{.}}{2016}]%
        {wu2016google}
\bibfield{author}{\bibinfo{person}{Yonghui Wu}, \bibinfo{person}{Mike
  Schuster}, \bibinfo{person}{Zhifeng Chen}, \bibinfo{person}{Quoc~V Le},
  \bibinfo{person}{Mohammad Norouzi}, \bibinfo{person}{Wolfgang Macherey},
  \bibinfo{person}{Maxim Krikun}, \bibinfo{person}{Yuan Cao},
  \bibinfo{person}{Qin Gao}, \bibinfo{person}{Klaus Macherey}, {et~al\mbox{.}}}
  \bibinfo{year}{2016}\natexlab{}.
\newblock \showarticletitle{Google's neural machine translation system:
  Bridging the gap between human and machine translation}.
\newblock \bibinfo{journal}{\emph{arXiv preprint arXiv:1609.08144}}
  (\bibinfo{year}{2016}).
\newblock


\bibitem[\protect\citeauthoryear{Yadid and Yahav}{Yadid and Yahav}{2016}]%
        {yadid2016extracting}
\bibfield{author}{\bibinfo{person}{Shir Yadid} {and} \bibinfo{person}{Eran
  Yahav}.} \bibinfo{year}{2016}\natexlab{}.
\newblock \showarticletitle{Extracting code from programming tutorial videos}.
  In \bibinfo{booktitle}{\emph{Proceedings of the 2016 ACM International
  Symposium on New Ideas, New Paradigms, and Reflections on Programming and
  Software}}. \bibinfo{pages}{98--111}.
\newblock


\bibitem[\protect\citeauthoryear{Yakdan, Eschweiler, Gerhards-padilla, and
  Smith}{Yakdan et~al\mbox{.}}{2015}]%
        {Yakdan2015}
\bibfield{author}{\bibinfo{person}{Khaled Yakdan}, \bibinfo{person}{Sebastian
  Eschweiler}, \bibinfo{person}{Elmar Gerhards-padilla}, {and}
  \bibinfo{person}{Matthew Smith}.} \bibinfo{year}{2015}\natexlab{}.
\newblock \showarticletitle{{No More Gotos: Decompilation Using
  Pattern-Independent Control-Flow Structuring and Semantics-Preserving
  Transformations}}. In \bibinfo{booktitle}{\emph{Network and Distributed
  System Security}}. \bibinfo{pages}{8--11}.
\newblock
\showISBNx{189156238X}


\bibitem[\protect\citeauthoryear{Yang, Dai, Yang, Carbonell, Salakhutdinov, and
  Le}{Yang et~al\mbox{.}}{2019}]%
        {yang2019xlnet}
\bibfield{author}{\bibinfo{person}{Zhilin Yang}, \bibinfo{person}{Zihang Dai},
  \bibinfo{person}{Yiming Yang}, \bibinfo{person}{Jaime Carbonell},
  \bibinfo{person}{Russ~R Salakhutdinov}, {and} \bibinfo{person}{Quoc~V Le}.}
  \bibinfo{year}{2019}\natexlab{}.
\newblock \showarticletitle{Xlnet: Generalized autoregressive pretraining for
  language understanding}. In \bibinfo{booktitle}{\emph{Advances in neural
  information processing systems}}. \bibinfo{pages}{5754--5764}.
\newblock


\bibitem[\protect\citeauthoryear{Yin and Neubig}{Yin and Neubig}{2017}]%
        {yin2017syntactic}
\bibfield{author}{\bibinfo{person}{Pengcheng Yin} {and} \bibinfo{person}{Graham
  Neubig}.} \bibinfo{year}{2017}\natexlab{}.
\newblock \showarticletitle{A syntactic neural model for general-purpose code
  generation}.
\newblock \bibinfo{journal}{\emph{arXiv preprint arXiv:1704.01696}}
  (\bibinfo{year}{2017}).
\newblock


\end{thebibliography}

\appendix

\section{Supplemental Material}

\end{document}